\documentclass[letterpaper]{article} 
\usepackage[preprint]{aaai2027}  
\usepackage[hyphens]{url}  
\usepackage{graphicx} 
\usepackage{natbib}  
\usepackage{caption} 
\usepackage{placeins}
\usepackage{algorithm}
\usepackage{algorithmic}
\usepackage{amsmath}
\usepackage{amssymb}
\usepackage{booktabs}
\usepackage{tabularx}
\usepackage{array}
\usepackage{booktabs}
\usepackage{multirow}
\usepackage{makecell}
\newcolumntype{Y}{>{\raggedright\arraybackslash}X}
\usepackage{newfloat}
\usepackage{listings}
\DeclareCaptionStyle{ruled}{labelfont=normalfont,labelsep=colon,strut=off} 
\floatstyle{ruled}
\newfloat{listing}{tb}{lst}{}
\floatname{listing}{Listing}

\usepackage{booktabs}

\title{Mode Connectivity Beyond Classifiers:\\Evidence from Generative and Contrastive Models}
\author{
    Chengzheyi Yao\textsuperscript{\rm 1},
    Yongzhao Zhang\textsuperscript{\rm 1},
    Yongding Tian\textsuperscript{\rm 2}
}
\affiliations{
    \textsuperscript{\rm 1}University of Electronic Science and Technology of China\\
    \textsuperscript{\rm 2}Delft University of Technology

}

\begin{document}

\maketitle

\begin{abstract}
The loss landscape of Deep Neural Networks (DNNs) exhibits highly complex and non-convex properties. Recent studies have revealed the phenomenon of mode connectivity, demonstrating that independently trained network modes can be connected via a continuous low-loss path. However, existing mode connectivity research is predominantly confined to classifier-based models, leaving it an open question whether similar geometric properties exist in modern complex models. In this paper, we extend the boundaries of mode connectivity to generative and contrastive domains (specifically DDPM and NanoCLIP). Addressing the unique architecture of DDPM and CLIP, we propose an architecture-aware connection building algorithm. Extensive empirical results demonstrate for the first time that we successfully discover mode connectivity between independently trained DDPM and NanoCLIP modes. Our work provides a novel perspective for understanding the geometric properties of the loss landscapes in modern generative and contrastive models.

\end{abstract}


\section{Introduction}



The loss functions of Deep Neural Networks (DNNs) are highly complex and non-convex, and their geometric properties have long been a subject of extensive research. Early visualizations of loss functions~\cite{landscape1, landscape2} supported the intuition that well-trained neural networks converge to isolated basins or local minima separated by high-loss barriers in the loss landscape. However, such visualizations are limited by their reliance on low-dimensional projections of the high-dimensional weight space. Further research suggests the basin model is incomplete.

Mode connectivity emerges by asking whether two independently trained neural networks can be connected by a continuous low-loss path in weight space. The endpoints of such a path are commonly referred to as \emph{modes}, i.e., well-trained models that attain low loss. Establishing mode connectivity is significant because traditional loss landscape visualizations typically capture low-dimensional slices and therefore fail to reveal the high-dimensional low-loss regions. In contrast, mode connectivity provides a tool for uncovering deeper geometric and topological properties of the loss landscape beyond the isolated-basin picture.

The study of mode connectivity has both theoretical and empirical roots. Singular learning theory~\cite{SLT} provided an algebraic-geometric perspective on this phenomenon: neural networks are singular statistical models, and the degeneracy of their Fisher information matrices~\cite{FisherMatrix} naturally gives rise to non-isolated parameter configurations and continuous low-loss subspaces. Empirical evidence for mode connectivity was later established. FGE~\cite{FGE} showed that independently trained modes can be connected by simple curves such as polygonal chains or Bezier curves. AutoNEB~\cite{AutoNEB} iteratively bends linear interpolations into low-loss paths by optimizing intermediate pivots. More recently, LLPF~\cite{LLPF} proposed a layer-wise algorithm, improving the reliability of connecting independently trained modes across a broader set of neural architectures.

Despite these advances, existing empirical evidence for mode connectivity remains largely confined to classifier-based settings, as shown in Table~\ref{tab:comparison}. Although these works cover multiple network architectures, the underlying learning paradigm is still relatively simple: unimodal supervised classification on small-scale CIFAR datasets. Whether mode connectivity exists in modern non-classifier architectures, such as generative and contrastive models, remains largely unexplored.



In this paper, we extend the scope of mode connectivity from classifiers to Denoising Diffusion Probabilistic Models (DDPM)~\cite{ddpm} on Flowers102~\cite{flowers102} and NanoCLIP~\cite{nanoclip} on Flickr30k~\cite{flickr30k}. These models introduce new challenges absent from classifiers: (1) DDPM uses a U-Net architecture with skip connections, while NanoCLIP relies on cross-modal alignment between image and text encoders. All previous methods~\cite{LLPF,layerwise} are not compatible with these models; and (2) these models have higher-dimensional parameter spaces than the classifiers, resulting in larger parameter-space distances between two modes. It is difficult to identify an effective optimization direction towards low-loss regions.


Our success in these models is primarily attributed to two key designs in two-stage mode connectivity pipeline~\cite{LLPF}: (1) in \textit{Weighted Movement} stage, we develop a dataflow-based layer moving strategy that determines the moving order from the computational dependencies of each architecture, building on the layer-wise methods; and (2) in \textit{Training Refinement} stage, we first identify that the choice of optimizer is important for higher-dimensional models. Thus, we employ the Adam optimizer~\cite{adamw} instead of Stochastic Gradient Descent (SGD), on which previous methods mainly rely.

Our main contributions are summarized as follows:
\begin{itemize}
\item To the best of our knowledge, we are the first to extend the scope of mode connectivity from conventional image classifiers to representative non-classifier models, including contrastive and generative models.
\item We provide empirical evidence that the mode connectivity property holds for DDPM on the Flowers102 dataset and for NanoCLIP on the Flickr30K dataset.
\end{itemize}

\section{Related Work}
\label{sec:Related_Work}

In this section, we present prior mode connectivity studies from three distinct paradigms.

\subsection{Theoretical Perspectives of Mode Connectivity}

Singular learning theory (SLT)~\cite{SLT} provides an important theoretical perspective for mode connectivity. In classical regular statistical models, different parameter values typically correspond to locally distinguishable predictive distributions, and the Fisher information matrix is non-degenerate around an optimum. Neural networks, however, are singular statistical models: the parameter-to-function map is generally non-injective due to overparameterization, permutation symmetries~\cite{LMCper1}, scaling invariances~\cite{scaling}, and redundant representations~\cite{redundant}. Consequently, the Fisher information matrix may become degenerate near well-trained solutions, and the set of parameters realizing the same or similar functions can form continuous singular structures rather than isolated points.

From the viewpoint of SLT, the existence of flat directions and low-loss subspaces is therefore a natural consequence of the algebraic-geometric structure of neural network parameterizations. This theoretical perspective indicates that trained modes may belong to extended low-loss regions in parameter space. However, SLT does not directly provide algorithms for connecting two given independently trained modes. As a result, empirical methods remain necessary for testing whether a low-loss path exists in neural networks.

\subsection{Empirical Studies of Mode Connectivity}

The investigation of mode connectivity initially emerged from an empirical approach, AutoNEB~\cite{AutoNEB}, which iteratively refines a linear path into a low-loss region by inserting and optimizing a sequence of intermediate pivots. It constructed continuous low-loss paths in architectures like ResNets~\cite{resnet} and DenseNets~\cite{densenet}. Despite these advancements, previous works were unable to find a low-loss path for any arbitrary mode pairs. To address this limitation, subsequent work~\cite{LLPF} introduced a Low-Loss Path Finding (LLPF) algorithm based on layer-wise connectivity. By moving the network layer-by-layer, their method successfully expanded to a wider range of architectures~\cite{EfficientNet, mobilenet, regnet, shufflenet,DLA,CCT} as mentioned in Table~\ref{tab:comparison}.

However, all of these empirical studies share a common limitation: their explorations are strictly confined to classifier-based models. The mode connectivity of generative and contrastive models remains entirely unexplored, which is the primary focus of our work.

\begin{table}[t]
\centering

\small
\setlength{\tabcolsep}{3pt}
\renewcommand{\arraystretch}{1.15}
\begin{tabularx}{\columnwidth}{p{0.18\columnwidth} Y Y}
\toprule
\textbf{Method} & \textbf{Model architectures} & \textbf{Datasets} \\
\midrule
FGE
&
VGG, ResNet, WideResNet
&
MNIST, CIFAR-10, CIFAR-100
\\
\midrule
AutoNEB
&
Basic CNN, ResNet, DenseNet
&
CIFAR-10, CIFAR-100
\\
\midrule
LLPF
&
All above and EfficientNet, MobileNet, RegNet, ShuffleNet, DLA, CCT
&
MNIST, CIFAR-10, CIFAR-100, ImageNet10
\\
\midrule
Ours
&
DDPM, NanoCLIP
&
Flowers102, Flickr30k
\\
\bottomrule
\end{tabularx}
\caption{Comparison of evaluated model architectures and datasets in prior mode connectivity studies.}
\label{tab:comparison}
\end{table}

\subsection{Mode Connectivity and Linear Mode Connectivity}

Although linear mode connectivity appears more direct compared to the nonlinear path, the midpoint of linear interpolation often encounters a severe loss barrier~\cite{AutoNEB}. It is crucial to highlight the difference between mode connectivity and linear mode connectivity. 

For mode connectivity, two independently trained modes correspond to distant and unaligned points. Finding a continuous path between them is already difficult in such a high-dimensional space. Requiring the entire path to remain inside the low-loss region makes the problem even more restrictive. As a result, existing studies in this category are primarily driven by empirical methods and remain limited in number. \emph{Our work only investigates mode connectivity in independently trained modes.} 

Linear mode connectivity is commonly observed under two settings: models either share an early portion of their optimization trajectory or are aligned with respect to parameter-space symmetries. In the former setting, models branched from a common early-stage checkpoint can be connected by a low-loss linear path~\cite{LMCspa1,LMCspa2}. In the latter setting, permutation-based weight matching can transform independently trained modes into aligned parameterizations that admit nearly barrier-free linear interpolation~\cite{LMCper1, LMCper2,LMCMoE}. In a nutshell, linear mode connectivity typically needs shared early training trajectories or invasive weight permutations to align models, which can make it easier to find a connected path by linear interpolation.
\section{Terminology and Definitions}
\label{sec:terminology}

To provide a unified notation for describing mode connectivity, we first define the key terms used throughout this paper.


\paragraph{Problem Definition.}

Mode connectivity can be understood as the problem of determining whether two independently trained modes can be connected by a continuous path in low-loss regions.  

A \textit{mode} refers to a well-trained parameter point that attains low loss. In this paper, we focus on independently trained modes, where different modes are obtained from different random initializations without shared early training checkpoints or weight permutation.

Given a neural network architecture $\mathcal{M}$, let $\theta \in \mathbb{R}^{D}$ denote its trainable parameters, where $D$ is the total number of trainable parameters. For geometric discussions, we use $P$ to denote the point in weight space corresponding to $\theta$. Each point $P \in \mathbb{R}^{D}$ is associated with a loss value $\mathcal{L}(P,\mathcal{D})$ evaluated on dataset $\mathcal{D}$. Given a loss threshold $L_{\mathrm{thres}}$, we define the low-loss region in weight space as:

\begin{equation}
S_{L \le L_{\mathrm{thres}}} := \{
P \in \mathbb{R}^{D}
\mid
\mathcal{L}(P,\mathcal{D}) \le L_{\mathrm{thres}}
\}
\label{eq:low_loss_region}
\end{equation}

\begin{figure}[t]
\centering
\includegraphics[width=0.99\columnwidth]{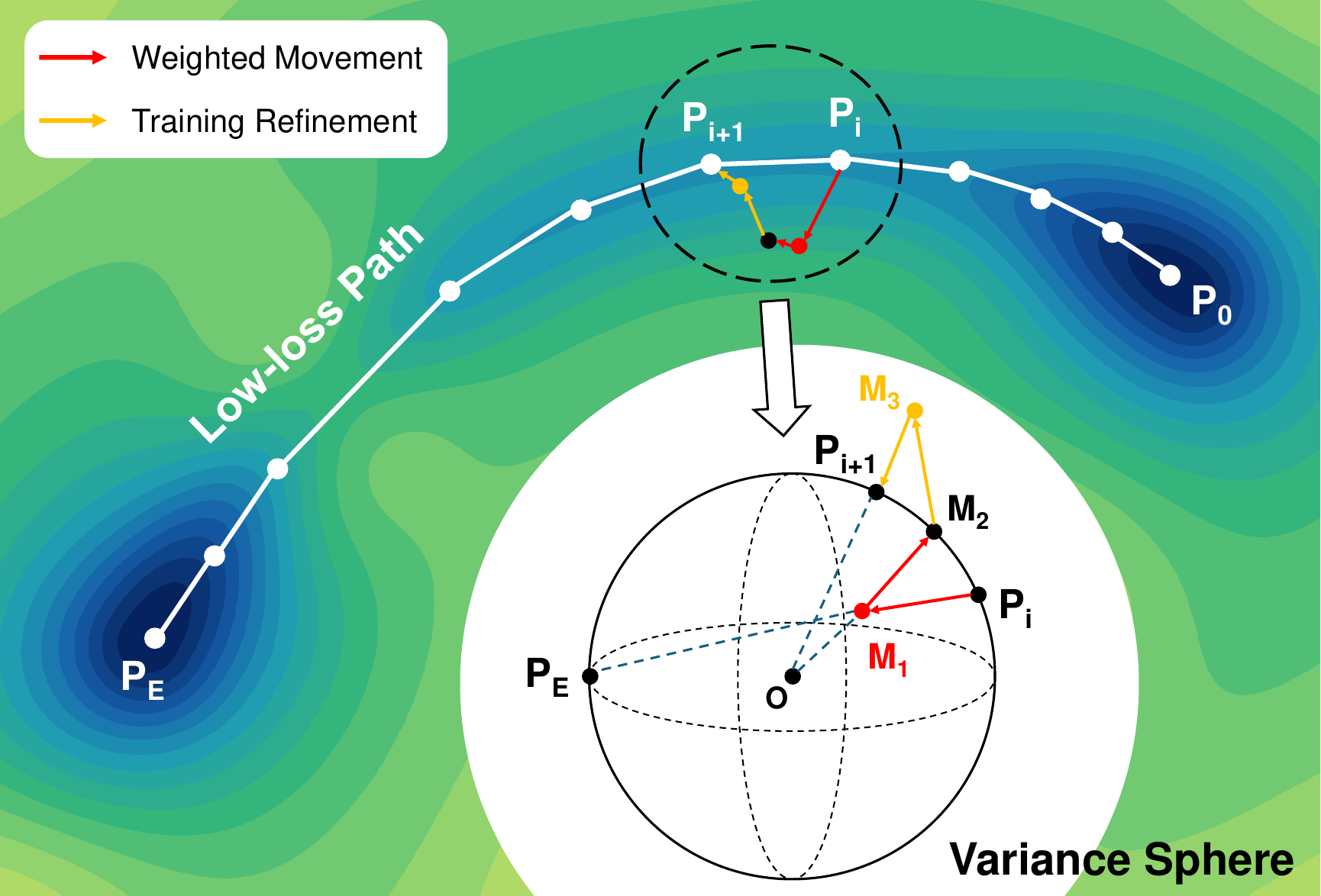}
\caption{Geometric illustration of architecture-aware connection building algorithm. The panel shows a slice of the loss function, where the endpoint modes are connected by a low-loss path. The variance sphere below zooms into a single iteration of the algorithm. Point labels follow the algorithm’s notation. The origin is used as the center of the variance sphere, assuming the mean is approximately zero (Equation~\ref{eq:approx_mean_var} and Equation~\ref{eq:variance_distance_relation}).}
\label{fig:variance_sphere}
\end{figure}

\paragraph{Variance Sphere.}
Neural networks are composed of multiple layers, which decompose the full parameter point $P$ as
$ P=[P^{(1)},P^{(2)},\dots,P^{(K)}]$
, where $P^{(k)}\in\mathbb{R}^{d_k}$ denotes the parameter vector of the $k$-th layer. 
Since the parameters of each layer can be viewed as a distribution of real values, we use $\mathrm{Mean}(P^{(k)})$ and $\mathrm{Var}(P^{(k)})$ to denote the empirical mean and variance of the $k$-th layer, respectively.

Based on the layer-wise variance, we define the \textit{Variance Sphere} as the set of layer parameters with a fixed variance value. For the $k$-th layer and a given variance $v$, the variance sphere is defined as:
\begin{equation}
S_{\mathrm{var}}^{(k)}(v):= \{
P^{(k)} \in \mathbb{R}^{d_k}
\mid
\mathrm{Var}(P^{(k)})=v \}
\label{eq:variance_sphere}
\end{equation}

For independently trained modes obtained under the same model architecture, their layer-wise mean and variance are expected to be close. This is because common initialization schemes for linear, convolutional, and transformer layers usually sample weights from zero-mean distributions. Their variance is determined by the layer size and computing method \cite{var1, var2, var3}. Therefore, two independent initializations of the same architecture satisfy:

\begin{equation}
\theta_0, \theta_0' = \mathrm{Init}(\mathcal{M})
\Longrightarrow
\mathrm{Var}(\theta_0) = \mathrm{Var}(\theta_0')
\label{eq:init_same_variance}
\end{equation}

As shown by the results in Appendix~\ref{appendix:meanvar}, Figure~\ref{fig:meanvar}, independently trained modes trained with identical hyperparameters on the same dataset tend to lie on nearby variance spheres, while the layer-wise means of the trained parameters typically remain near zero. Thus, for the $k$-th layer, we adopt the approximation:
\begin{equation}
\mathrm{Mean}(P^{(k)})\approx 0 \quad \text{and} \quad \mathrm{Var}(\theta_n^{k}) \approx \mathrm{Var}(\theta_n'^{k})
\label{eq:approx_mean_var}
\end{equation}
Under this approximation, the variance of a layer is approximately proportional to the squared Euclidean distance from the corresponding parameter point to the origin in $\mathbb{R}^{d_k}$:
\begin{equation}
\left\| \overrightarrow{OP^{(k)}} \right\| ^2
\propto
\mathrm{Var}(P^{(k)})
\label{eq:variance_distance_relation}
\end{equation}
Therefore, fixing the variance of a layer approximately fixes its distance to the origin. This explains why $S_{\mathrm{var}}^{(k)}(v)$ can be interpreted geometrically as a high-dimensional sphere in the layer-wise parameter space.

\section{Architecture-aware Connection Building Algorithm}
\label{sec:algorithm}

\begin{figure*}[t]
\centering
\includegraphics[width=1\textwidth]{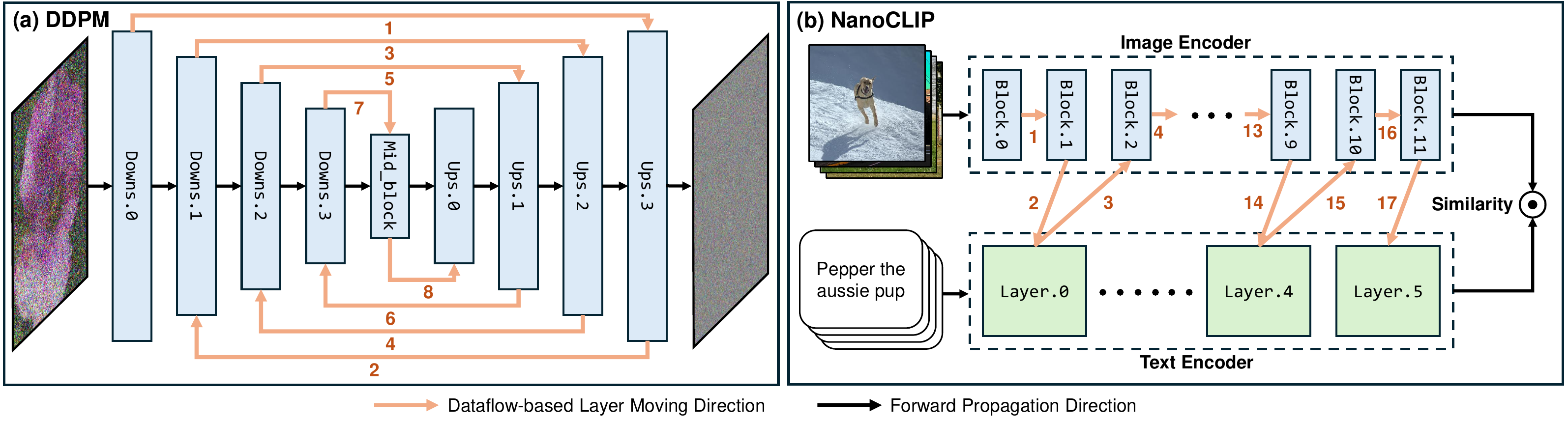}
\caption{
Dataflow-based layer moving strategies for DDPM and NanoCLIP. \textit{Left}: DDPM layers are added to the moving sequence following a symmetrical head-to-tail schedule. \textit{Right}: NanoCLIP layers are added through an alternating visual-textual schedule. The Layer Moving Direction indicates the order in which layer groups are introduced into layer moving sequence $\mathcal{U}_i$.}
\label{fig:moving_strategy}
\end{figure*}

\begin{algorithm}[t]
\caption{Architecture-aware Connection Building
Algorithm}
\label{alg:aacb}
\begin{algorithmic}[1]

\small

\REQUIRE Endpoint modes $P_0$ and $P_E$, dataset $\mathcal{D}$, step-size hyperparameters $(\alpha,\beta,\gamma)$, training hyperparameters $\xi$, layer moving sequence $\mathcal{U}$, total iterations $T_{total}$.

\STATE $P_i = P_0, \mathcal{P} .append( P_i )$

\FOR{$i = 0$ \TO $T_{total}-1$}
\STATE $step = \alpha \left| \overrightarrow {P_iP_E} \right| + \beta \left| \overset{\frown}{P_0P_E} \right| + \gamma$

\STATE $M_1 = P_i + step \cdot \Pi_{\mathcal{U}_i}\!\left(\overrightarrow{P_iP_E}\right)$

\STATE $M_2 = \mathrm{VarianceCorrection}(M_1, S_{Var})$

\STATE $M_3 =\mathrm{Train}(M_2, \xi, \mathcal{D})$.

\STATE $P_{i+1} = \mathrm{VarianceCorrection}(M_3, S_{Var})$

\STATE $\mathcal{P}.append( {P_{i+1}} )$.
\ENDFOR
\RETURN $\mathcal{P}$

\STATE
\STATE \textbf{Function} $\mathrm{VarianceCorrection}(Q, S_{Var})$
\FOR{$W = Q^{(k)}$ in $Q$}
    \STATE $\bar{W} = \frac{1}{n}\sum_{j=1}^{n} W[j]$
    \STATE $\sigma_W^2 \gets \frac{1}{n}\sum_{j=1}^{n}(W[j]-\bar{W})^2$
    \FOR{$j=1$ to len($W$)}
        \STATE $W'[j] = \bar{W}
        + \sqrt{\frac{v}{\sigma_W^2}}\bigl(W[j]-\bar{W}\bigr)$
    \ENDFOR
    \STATE $Q'^{(k)} = W'$
\ENDFOR
\RETURN $Q'$

\end{algorithmic}
\end{algorithm}

This section presents the proposed architecture-aware connection building algorithm. We first introduce two-stage connection building framework of our algorithm and then detail two key designs in \textit{Weighted Movement} and \textit{Training Refinement} as shown in Figure~\ref{fig:variance_sphere}.

\subsection{Mode-to-mode Connection Building Framework}

The core objective of connection building is to iteratively transition from $P_0$ toward $P_E$ while ensuring all anchor points remain within the low-loss regions, as presented by the low-loss path in Figure~\ref{fig:variance_sphere}. The corresponding procedure is formalized in Algorithm~\ref{alg:aacb} and consists of two stages:

\paragraph{Weighted Movement ($P_i \rightarrow M_2$).} We first move the active layers of $P_i$ to obtain $M_1$ toward the destination $P_E$, controlled by $(\alpha,\beta,\gamma)$ and $\mathcal{U}_i$. Because averaging independently trained weights reduces their layer-wise variance, it can hinder subsequent training process~\cite{varianceproblem}. Variance correction rescales $M_1$ to $M_2$, which lies on the reference variance sphere of $P_0$ as shown in Figure~\ref{fig:variance_sphere}.

\paragraph{Training Refinement ($M_2 \rightarrow P_{i+1}$).} Restoring the layer-wise variance does not by itself guarantee that $M_2$ lies in low-loss regions, so we then train $M_2$ for several rounds, yielding $M_3$. The variance correction is applied once more to define the next anchor $P_{i+1}$. Repeating this procedure for $T_{total}$ iterations generates a sequence from $P_0$ to $P_T \approx P_E$, with consecutive points forming the desired low-loss path.

\subsection{Dataflow-based Layer Moving Strategy}

The most critical hyperparameter of \textit{Weighted Movement} is the layer moving sequence $\mathcal{U}_i$, which decides the subset of layers needed to move at a stage. Moving all layers at once failed due to complex architectures of DDPM and NanoCLIP. To solve this, we design dataflow-based layer moving strategy that follows two rules: (1) layers should be processed in the direction of data flow according to model architectures; and (2) attention modules should be processed individually before proceeding to subsequent layers. Figure~\ref{fig:moving_strategy} contrasts the ordinary forward propagation directions with our moving strategy for DDPM and NanoCLIP.

\paragraph{Symmetrical Strategy for DDPM.}

For DDPM, the network typically adopts a U-Net topology characterized by symmetrical contracting and expanding paths linked via skip connections. Since corresponding downsampling and upsampling blocks handle features at identical spatial resolutions, moving them separately may disrupt the denoising trajectory. To address this, we introduce a symmetrical head-to-tail strategy where corresponding layers from the encoder and decoder are added to $\mathcal{U}_i$ sequentially, progressing from the outermost layers to the innermost layers. By binding these structurally coupled layers into unified transformation units, the algorithm preserves the integrity of the high-to-low feature hierarchy, enabling stable mode connectivity across generative manifolds.

\paragraph{Alternating Strategy for NanoCLIP.}

The NanoCLIP architecture utilizes a dual-encoder paradigm where visual and textual features are projected into a shared embedding space via contrastive learning. To maintain the semantic alignment between modalities during path construction, we propose an alternating strategy. Specifically, two visual encoder blocks are added to the $\mathcal{U}_i$, followed by a single layer of the text encoder, and this cross-modal coordination repeats sequentially across all available layers. This strategy ensures that the visual representation space and textual representation space deform symmetrically, preventing the collapse of contrastive alignment along the path.

\begin{figure*}[t]
    \centering
    \includegraphics[width=0.99\textwidth]{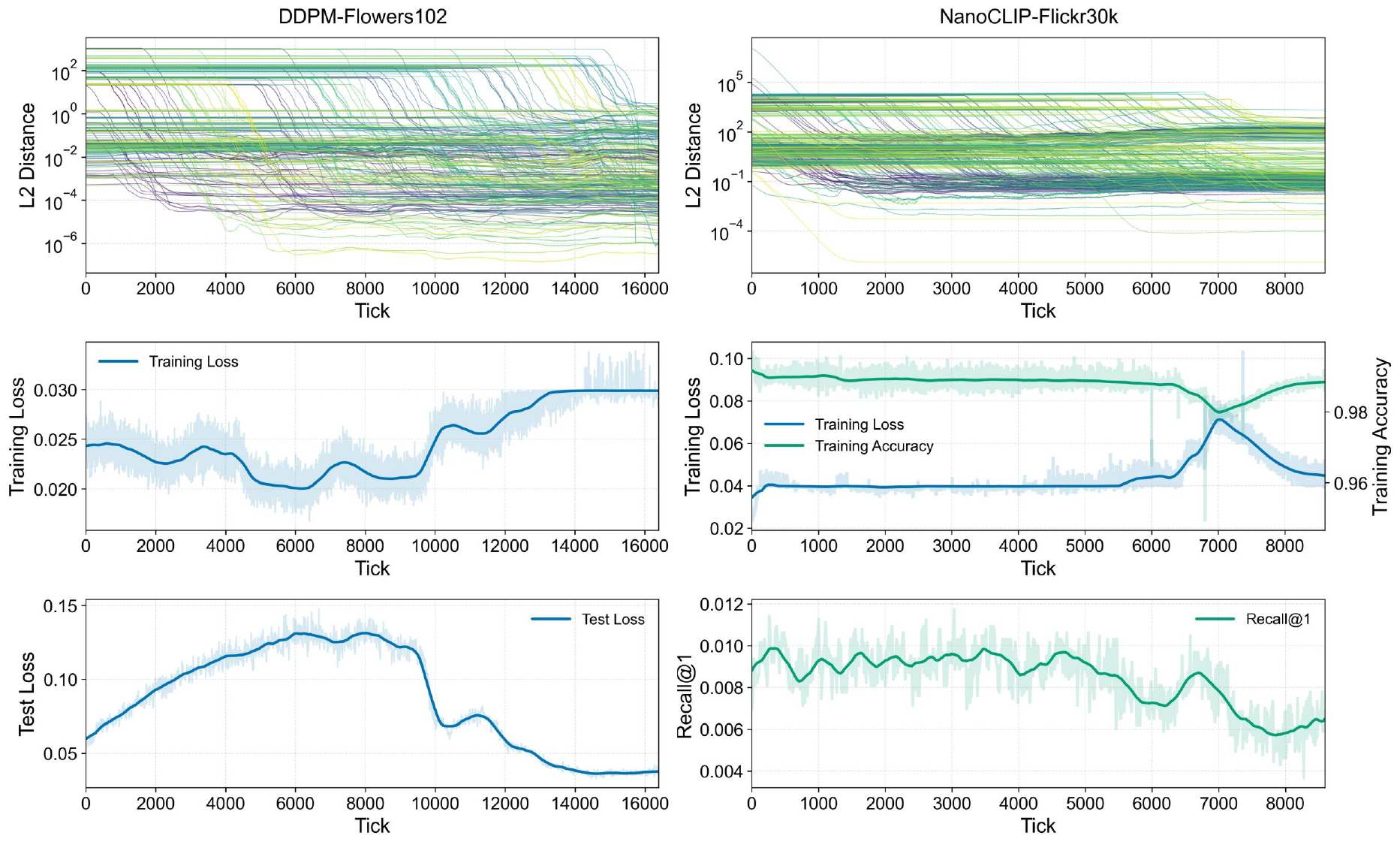}
    \caption{Results of the architecture-aware connection building algorithm on DDPM (first column) and NanoCLIP (second column). The first row shows the layer-wise $L_2$ distance from the current point $P_i$ to the end mode $P_E$,  and the legend of layers is omitted due to excessive quantity. The second and third rows respectively report the training and test metrics along the path. The curves show the smoothed trends, and the shaded regions indicate local fluctuation envelopes of the raw metrics. Due to the learning objectives of the models, DDPM is evaluated only with training and test losses, whereas NanoCLIP is evaluated without a test loss.}
    \label{fig:m2m_results}
\end{figure*}

\subsection{Adam-based Refinement for Distant Modes}

The purpose of \textit{Training Refinement} is to return the variance-corrected point $M_2$ to the low-loss region after each weighted movement. In the classifier settings, the movements are typically short and localized. Under this condition, a SGD optimizer is often sufficient to identify a nearby descent direction and recover low loss. The DDPM and NanoCLIP models operate in substantially higher-dimensional parameter spaces than the classifier-based models. As a result, independently trained checkpoints exhibit larger parameter distances, including larger layer-wise separations. Empirically, we find that the single global learning-rate scaling of SGD becomes unreliable in this regime: SGD may reduce the loss for small endpoint separations, but frequently fails to restore low loss when the checkpoint separation is large.

We address this limitation by implementing training refinement with Adam. Its first and second moment estimates provide coordinate-wise adaptive scaling, which is useful when different modules exhibit substantially different gradient magnitudes. The momentum state also accumulates a stable refinement direction across successive updates. During each refinement phase, Adam updates the parameters once the training loss falls below the threshold or the maximum number of refinement rounds is reached. The optimizer ablation in Figure~\ref{fig:ablation} supports this design: replacing Adam with SGD while retaining all other components prevents the refinement step from maintaining a low-loss trajectory.
\section{Experimental Evaluation}
\label{sec:Evaluation}

To evaluate the proposed architecture-aware connection building algorithm, we consider two representative models: DDPM on Flowers102 and NanoCLIP on Flickr30k. For each model, we independently train two modes using different random seeds and subsequently apply our method to construct a low-loss path between them. All experiments follow the architecture-aware layer-moving schedules and configurations described in Appendix~\ref{appendix:hyperparameters}, Table~\ref{tab:Hyperparameters}.


\begin{figure*}[t]
    \centering
    \includegraphics[width=0.99\textwidth]{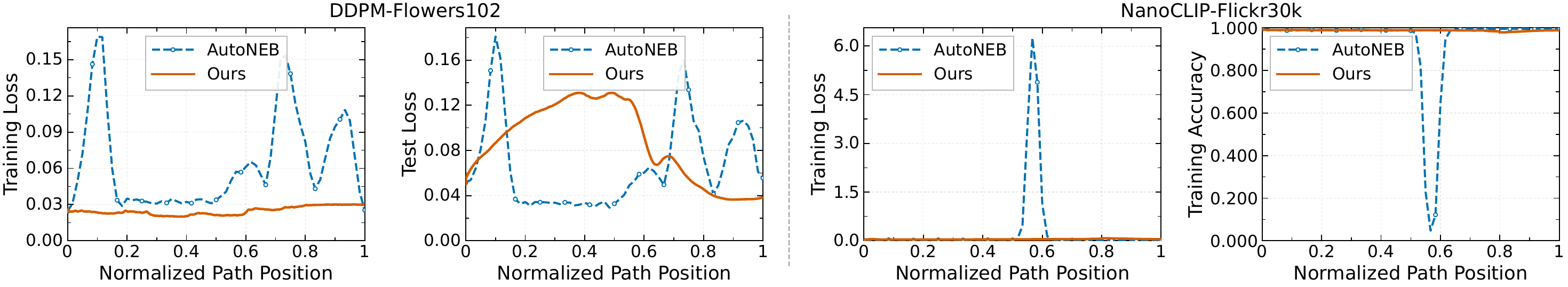}
    \caption{Comparison of AutoNEB and our method along constructed paths on DDPM and NanoCLIP. From left to right, the four panels report DDPM training loss, DDPM test loss, NanoCLIP training loss, and NanoCLIP training accuracy, respectively. The horizontal axis denotes the normalized path position, where 0 and 1 correspond to the starting and end modes. The dots on the dashed line are the pivots optimized by AutoNEB. Between adjacent pivots, we also evaluate the metrics of interpolation points along each segment. For visual clarity, the densely sampled curves of our method are smoothed.}
    \label{fig:comparison}
\end{figure*}

\paragraph{Overall Performance.}

Figure~\ref{fig:m2m_results} evaluates the validity of the constructed paths from three complementary perspectives: (1) for the first row, the layer-wise $L_2$ distance between the current point $P_i$ and the end mode $P_E$ should decrease progressively and eventually approach the destination. Small deviations are acceptable because each stage only approximates the ideal movement; (2) for the second row, the training loss should remain within the low-loss threshold throughout the connection process; and (3) for the third row, the test evaluation metrics at the end of the path should converge to those of the starting mode $P_0$, indicating that the constructed path not only reaches the target parameter region but also recovers its generalization performance.

The first row of Figure~\ref{fig:m2m_results} shows the evolution of the layer-wise $L_2$ distances. For both DDPM and NanoCLIP, the distances decrease stage by stage following the predefined layer moving strategy, demonstrating that the proposed algorithm consistently moves the current point toward the end mode. Although the final distances of several layers do not converge exactly to zero, they are reduced by at least one order of magnitude compared with their initial values. This behavior is expected because different layers contain substantially different numbers of parameters, leading to different distance scales. Consequently, the relative reduction from the initial distance more accurately reflects the convergence behavior.

For DDPM, the denoising training loss remains below 0.04 throughout the entire connection process, while the testing loss eventually converges to approximately 0.04, which is even lower than the starting mode. For NanoCLIP, the contrastive training loss remains below 0.09 during optimization, while both the training accuracy and $Recall@1$ stay relatively stable. These results demonstrate that the proposed algorithm is capable of constructing low-loss paths while preserving their generation capability.

Because the procedure produces a discrete sequence of checkpoints, we further verify that the line segments between consecutive checkpoints do not introduce hidden loss barriers. As detailed in Appendix~\ref{appendix:continuity}, dense linear interpolation along the NanoCLIP trajectory reveals no observable barrier, providing empirical evidence that the path remains within the low-loss region.

\paragraph{Comparison Study.}

Among a limited number of mode connectivity methods, FGE, AutoNEB and LLPF are the relevant baselines for comparison. However, the released implementation of FGE contains an error, and its procedure does not necessarily yield a valid low-loss path~\cite{LLPF}. Our method is based on the layer-wise paradigm of LLPF, so the comparative results with LLPF can actually refer to the ablation results combined with forward order and SGD. We therefore select AutoNEB as the primary baseline. For a controlled comparison, we follow the original AutoNEB training configuration and adapt its pipeline to accommodate DDPM and NanoCLIP as shown in Appendix~\ref{appendix:autoneb_config}, Table~\ref{tab:AutoNEBHyperparameters}.

Figure~\ref{fig:comparison} compares AutoNEB with our method along the constructed paths. On DDPM, the maximum training and test losses obtained by AutoNEB are 0.17 and 0.18, respectively, whereas our method limits them to 0.03 and 0.15. The difference is substantially larger on NanoCLIP: AutoNEB reaches a maximum training loss of 6.69 and a minimum training accuracy of only 0.05, while our method achieves a maximum training loss of 0.09 and maintains a minimum training accuracy of 0.95. These results show that AutoNEB fails to construct a low-loss path for generative and contrastive models. In contrast, our method maintains consistently low training loss throughout the path.

\begin{figure}[t]
    \centering
    \includegraphics[width=0.99\columnwidth]{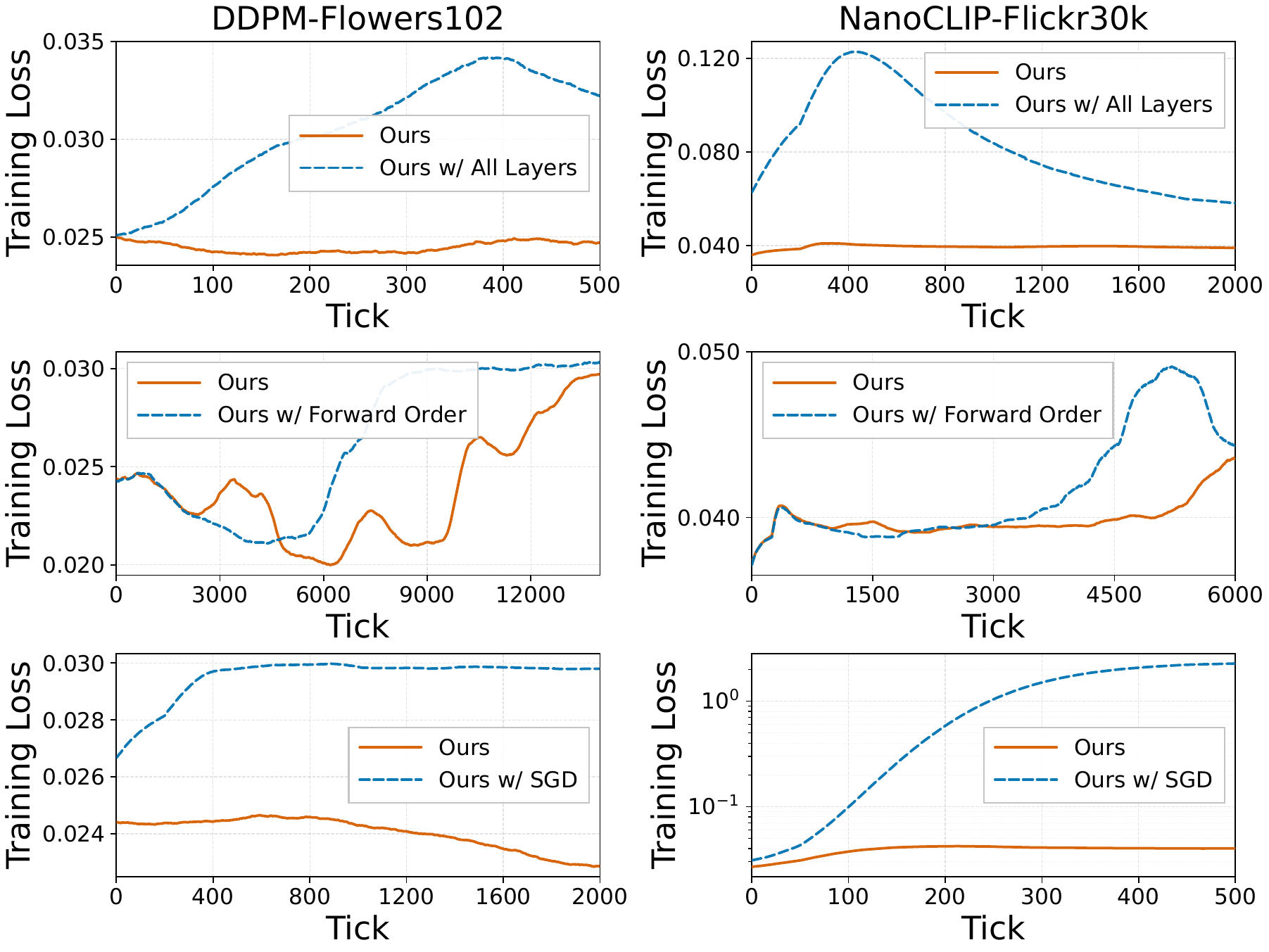}
    \caption{Ablation study of the proposed architecture-aware connection building algorithm on DDPM and NanoCLIP. The left and right columns report the results on DDPM and NanoCLIP, respectively. The first row compares the proposed dataflow-based layer moving strategy with simultaneous movement of all layers. The second row compares the proposed strategy with the forward propagation order. The third row replaces Adam with SGD while keeping the remaining components unchanged. All curves are smoothed.}
    \label{fig:ablation}
\end{figure}

\paragraph{Ablation Study.}

We examine two key designs of our method: dataflow-based layer moving strategy and Adam-based training refinement. Figure~\ref{fig:ablation} compares the complete method with three variants: \emph{Ours w/ All Layers}, which moves all model layers simultaneously; \emph{Ours w/ Forward Order}, which retains staged movement but activates layers according to the forward propagation order; and \emph{Ours w/ SGD}, which replaces Adam with SGD. All remaining components and hyperparameters are kept unchanged unless otherwise specified. It is worth noting that we did not construct a complete path in all experiments, as sufficient differences could already be observed within limited ticks.

The first row of Figure~\ref{fig:ablation}  investigates whether the dataflow-based layer moving strategy can be replaced by moving all layers simultaneously. On DDPM, the all-layers variant yields an average training loss of 0.031, compared with 0.025 of the complete method. On NanoCLIP, the average training loss of the variant is more than twice that of the complete method. These results indicate that moving all layers increases the optimization difficulty, whereas the proposed strategy more effectively preserves a low-loss trajectory.

The second row compares the proposed strategy with the forward propagation order over the complete trajectories. Our strategy outperforms the forward-order variant on both DDPM and NanoCLIP, reducing the average training loss by 0.002 in each case. This indicates that our dataflow-based order provides a more effective movement schedule than computational order for movement in weight space.

The third row evaluates the optimizer used during training refinement. On DDPM, the complete method achieves an average training loss of 0.024, compared with 0.030 for the SGD variant. The difference is more pronounced on NanoCLIP, where its training loss can increase by two orders of magnitude. These results indicate that the coordinate-wise adaptive updates of Adam improve refinement stability when different models exhibit substantially different gradient magnitudes. Overall, the ablation results show that Adam-based refinement and dataflow-based layer moving strategy provide complementary benefits for preserving low training loss along the constructed paths.
\section{Discussion}
\label{sec:Discussion}

Beyond validating the effectiveness of the proposed method, our empirical results also reveal several observations.

\begin{figure}[t]
    \centering
    \includegraphics[width=0.99\columnwidth]{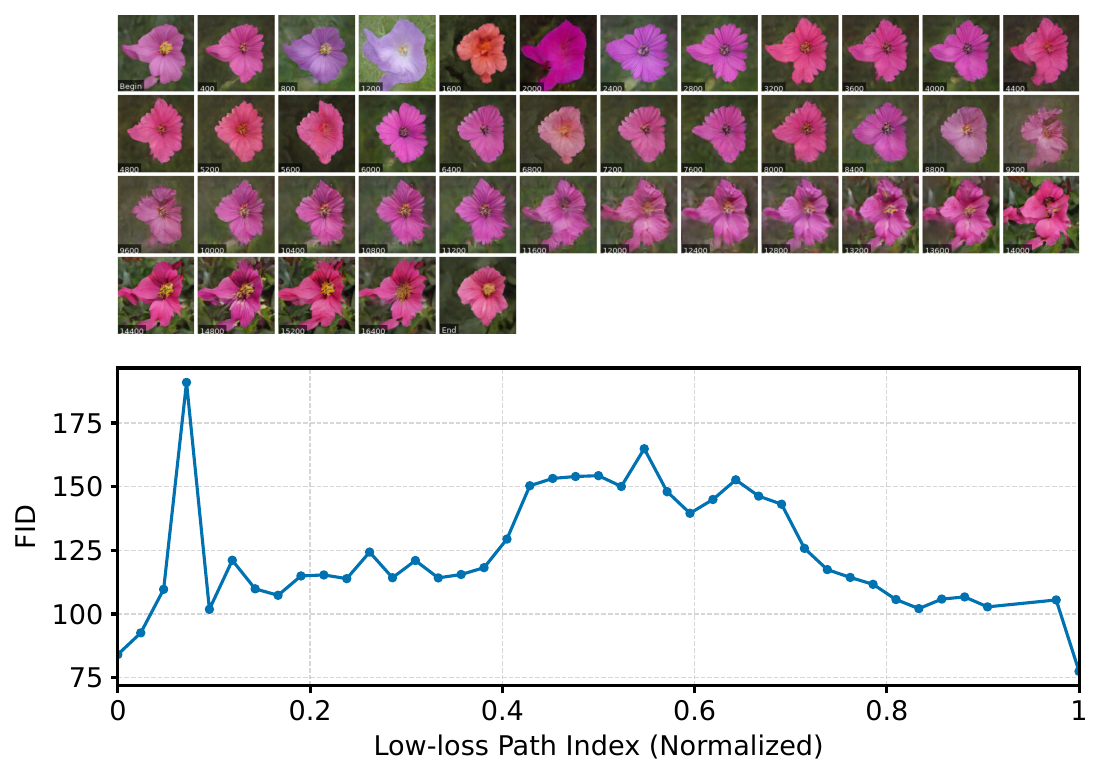}
    \caption{Generation quality evaluation of checkpoints along the constructed low-loss path. \textit{Top}: representative synthesized images sampled from identical initial Gaussian noise. \textit{Bottom}: the corresponding FID computed across the 41 sampled checkpoints.
}
    \label{fig:ddpm_generate}
\end{figure}

\paragraph{Low training loss does not imply high generation quality.} Although mode connectivity is defined primarily by low training loss, examining the image generation capability of intermediate DDPM models provides a complementary assessment. We uniformly sample 41 checkpoints along the path, including both endpoints. Each checkpoint generates 1000 images under an identical sampling configuration and random seed, so corresponding samples differ only in model parameters. Figure~\ref{fig:ddpm_generate} shows one representative image from each checkpoint together with its Fréchet Inception Distance (FID).
Quantitatively, the intermediate checkpoint generally produces higher FID values than the endpoints, while the training loss of the checkpoints remains basically the same as that of the endpoint. This behavior is expected because the intermediate models are optimized to satisfy the low-loss constraint rather than to minimize FID directly. The results may also indicate that a low-loss region is not necessarily a high-quality generative region.

\paragraph{Endpoint parameter proximity does not imply a flat basin.} 

Prior studies describe well-trained minima as lying within locally flat manifolds. Under this interpretation, once a constructed trajectory approaches the end mode, the remaining neighborhood should be sufficiently flat that direct interpolation introduces little additional loss. We test this expectation around the DDPM endpoint.
As shown in the left panel of Figure~\ref{fig:ddpm_barrier} in Appendix~\ref{appendix:barrier}, the checkpoint at tick 16400 is already strongly aligned with the end mode: layer-wise cosine similarities are concentrated near one across all DDPM modules, indicating that the checkpoint is close to the endpoint in weight space. Nevertheless, the right panel shows that linear interpolation between them creates a maximum training loss of $0.038$, compared with a maximum of $0.034$ along the previous low-loss path. One possible interpretation is that the low-loss region near the end mode is highly anisotropic: even near a trained mode, the local geometry of foundation-model loss landscapes may be more intricate than the conventional picture of a uniformly flat basin.




\section{Conclusion}
\label{sec:Conclusion}

We investigated whether mode connectivity extends beyond classifiers to independently trained models with generative and contrastive objectives. To address the architectural and optimization challenges posed by DDPM and NanoCLIP, we proposed an architecture-aware connection building algorithm that combines dataflow-based layer movement with Adam-based refinement. Experiments on DDPM trained on Flowers102 and NanoCLIP trained on Flickr30k show that the proposed method constructs continuous low-loss paths while progressively reducing layer-wise parameter distances and maintaining low training loss. Overall, our work is the first to extend mode connectivity to generative and contrastive domains and motivate further studies of loss-landscape geometry in modern non-classifier models.

\bibliography{aaai2027}

@book{SLT,
author = {Watanabe, Sumio},
title = {Algebraic Geometry and Statistical Learning Theory},
year = {2009},
isbn = {0521864674},
publisher = {Cambridge University Press},
address = {USA}
}

@inproceedings{
layerwise,
title={Layer-wise linear mode connectivity},
author={Linara Adilova and Maksym Andriushchenko and Michael Kamp and Asja Fischer and Martin Jaggi},
booktitle={The Twelfth International Conference on Learning Representations},
year={2024},
url={https://openreview.net/forum?id=LfmZh91tDI}
}

@inproceedings{FGE,
author = {Garipov, Timur and Izmailov, Pavel and Podoprikhin, Dmitrii and Vetrov, Dmitry and Wilson, Andrew Gordon},
title = {Loss surfaces, mode connectivity, and fast ensembling of DNNs},
year = {2018},
publisher = {Curran Associates Inc.},
address = {Red Hook, NY, USA},
booktitle = {Proceedings of the 32nd International Conference on Neural Information Processing Systems},
pages = {8803–8812},
numpages = {10},
location = {Montr\'{e}al, Canada},
series = {NIPS'18}
}

@inproceedings{
LLPF,
title={Connecting Independently Trained Modes via Layer-Wise Connectivity},
author={Yongding Tian and Zaid Al-Ars and Maksim Kitsak and H Peter Hofstee},
booktitle={Forty-third International Conference on Machine Learning},
year={2026},
url={https://openreview.net/forum?id=4VOTzpH9MO}
}

@InProceedings{AutoNEB,
  title = 	 {Essentially No Barriers in Neural Network Energy Landscape},
  author =       {Draxler, Felix and Veschgini, Kambis and Salmhofer, Manfred and Hamprecht, Fred},
  booktitle = 	 {Proceedings of the 35th International Conference on Machine Learning},
  pages = 	 {1309--1318},
  year = 	 {2018},
  editor = 	 {Dy, Jennifer and Krause, Andreas},
  volume = 	 {80},
  series = 	 {Proceedings of Machine Learning Research},
  month = 	 {10--15 Jul},
  publisher =    {PMLR},
  url = 	 {https://proceedings.mlr.press/v80/draxler18a.html}
}

@inproceedings{
LMCMoE,
title={On Linear Mode Connectivity of Mixture-of-Experts Architectures},
author={Viet-Hoang Tran and Van-Hoan Trinh and Khanh Vinh Bui and Tan Minh Nguyen},
booktitle={The Thirty-ninth Annual Conference on Neural Information Processing Systems},
year={2025},
url={https://openreview.net/forum?id=RF3miSqdXa}
}

@InProceedings{LMCspa1,
  title = 	 {Linear Mode Connectivity and the Lottery Ticket Hypothesis},
  author =       {Frankle, Jonathan and Dziugaite, Gintare Karolina and Roy, Daniel and Carbin, Michael},
  booktitle = 	 {Proceedings of the 37th International Conference on Machine Learning},
  pages = 	 {3259--3269},
  year = 	 {2020},
  editor = 	 {III, Hal Daumé and Singh, Aarti},
  volume = 	 {119},
  series = 	 {Proceedings of Machine Learning Research},
  month = 	 {13--18 Jul},
  publisher =    {PMLR},
  url = 	 {https://proceedings.mlr.press/v119/frankle20a.html}
}

@inproceedings{LMCspa2,
author = {Fort, Stanislav and Dziugaite, Gintare Karolina and Paul, Mansheej and Kharaghani, Sepideh and Roy, Daniel M. and Ganguli, Surya},
title = {Deep learning versus kernel learning: an empirical study of loss landscape geometry and the time evolution of the neural tangent kernel},
year = {2020},
isbn = {9781713829546},
publisher = {Curran Associates Inc.},
address = {Red Hook, NY, USA},
booktitle = {Proceedings of the 34th International Conference on Neural Information Processing Systems},
articleno = {491},
numpages = {12},
location = {Vancouver, BC, Canada},
series = {NIPS '20}
}

@inproceedings{
LMCper2,
title={Git Re-Basin: Merging Models modulo Permutation Symmetries},
author={Samuel Ainsworth and Jonathan Hayase and Siddhartha Srinivasa},
booktitle={The Eleventh International Conference on Learning Representations },
year={2023},
url={https://openreview.net/forum?id=CQsmMYmlP5T}
}

@inproceedings{
LMCper1,
title={The Role of Permutation Invariance in Linear Mode Connectivity of Neural Networks},
author={Rahim Entezari and Hanie Sedghi and Olga Saukh and Behnam Neyshabur},
booktitle={International Conference on Learning Representations},
year={2022},
url={https://openreview.net/forum?id=dNigytemkL}
}

@inproceedings{
landscape1,
title={Sharpness-aware Minimization for Efficiently Improving Generalization},
author={Pierre Foret and Ariel Kleiner and Hossein Mobahi and Behnam Neyshabur},
booktitle={International Conference on Learning Representations},
year={2021},
url={https://openreview.net/forum?id=6Tm1mposlrM}
}

@inproceedings{landscape2,
 author = {Li, Hao and Xu, Zheng and Taylor, Gavin and Studer, Christoph and Goldstein, Tom},
 booktitle = {Advances in Neural Information Processing Systems},
 editor = {S. Bengio and H. Wallach and H. Larochelle and K. Grauman and N. Cesa-Bianchi and R. Garnett},
 pages = {},
 publisher = {Curran Associates, Inc.},
 title = {Visualizing the Loss Landscape of Neural Nets},
 url = {https://proceedings.neurips.cc/paper_files/paper/2018/file/a41b3bb3e6b050b6c9067c67f663b915-Paper.pdf},
 volume = {31},
 year = {2018}
}

@proceedings{var1,
editor = {Genevieve B. Orr and Klaus-Robert M\"{u}ller},
title = {Neural Networks: Tricks of the Trade, this book is an outgrowth of a 1996 NIPS workshop},
year = {1998},
isbn = {3540653112},
publisher = {Springer-Verlag},
address = {Berlin, Heidelberg}
}

@InProceedings{var2,
  title = 	 {Understanding the difficulty of training deep feedforward neural networks},
  author = 	 {Glorot, Xavier and Bengio, Yoshua},
  booktitle = 	 {Proceedings of the Thirteenth International Conference on Artificial Intelligence and Statistics},
  pages = 	 {249--256},
  year = 	 {2010},
  editor = 	 {Teh, Yee Whye and Titterington, Mike},
  volume = 	 {9},
  series = 	 {Proceedings of Machine Learning Research},
  address = 	 {Chia Laguna Resort, Sardinia, Italy},
  month = 	 {13--15 May},
  publisher =    {PMLR},
  url = 	 {https://proceedings.mlr.press/v9/glorot10a.html}
}

@INPROCEEDINGS{var3,
  author={He, Kaiming and Zhang, Xiangyu and Ren, Shaoqing and Sun, Jian},
  booktitle={2015 IEEE International Conference on Computer Vision (ICCV)}, 
  title={Delving Deep into Rectifiers: Surpassing Human-Level Performance on ImageNet Classification}, 
  year={2015},
  volume={},
  number={},
  pages={1026-1034},
  doi={10.1109/ICCV.2015.123}}

@misc{varianceproblem,
      title={Vanishing Variance Problem in Fully Decentralized Neural-Network Systems}, 
      author={Yongding Tian and Zaid Al-Ars and Maksim Kitsak and Peter Hofstee},
      year={2024},
      eprint={2404.04616},
      archivePrefix={arXiv},
      primaryClass={cs.LG},
      url={https://arxiv.org/abs/2404.04616}, 
}

@misc{nanoclip,
  author       = {Ali-bey, Amar},
  title        = {{nanoCLIP}: A Lightweight Text-to-Image Retrieval Model},
  year         = {2024},
  howpublished = {\url{https://github.com/amaralibey/nanoCLIP}},
  note         = {GitHub repository, accessed July 14, 2026}
}

@article{flickr30k,
  title={From image descriptions to visual denotations: New similarity metrics for semantic inference over event descriptions},
  author={Young, Peter and Lai, Alice and Hodosh, Micah and Hockenmaier, Julia},
  journal={Transactions of the Association for Computational Linguistics},
  volume={2},
  pages={67--78},
  year={2014},
  publisher={MIT Press}
}

@misc{ddpm,
      title={Denoising Diffusion Probabilistic Models}, 
      author={Jonathan Ho and Ajay Jain and Pieter Abbeel},
      year={2020},
      eprint={2006.11239},
      archivePrefix={arXiv},
      primaryClass={cs.LG},
      url={https://arxiv.org/abs/2006.11239}, 
}

@InProceedings{flowers102,
   author = "Nilsback, M-E. and Zisserman, A.",
   title = "Automated Flower Classification over a Large Number of Classes",
   booktitle = "Proceedings of the Indian Conference on Computer Vision, Graphics and Image Processing",
   year = "2008",
   month = "Dec"
}

@INPROCEEDINGS {mobilenet,
author = { Sandler, Mark and Howard, Andrew and Zhu, Menglong and Zhmoginov, Andrey and Chen, Liang-Chieh },
booktitle = { 2018 IEEE/CVF Conference on Computer Vision and Pattern Recognition (CVPR) },
title = {{ MobileNetV2: Inverted Residuals and Linear Bottlenecks }},
year = {2018},
volume = {},
ISSN = {},
pages = {4510-4520},
doi = {10.1109/CVPR.2018.00474},
url = {https://doi.ieeecomputersociety.org/10.1109/CVPR.2018.00474},
publisher = {IEEE Computer Society},
address = {Los Alamitos, CA, USA},
month =Jun}

@INPROCEEDINGS {shufflenet,
author = { Zhang, Xiangyu and Zhou, Xinyu and Lin, Mengxiao and Sun, Jian },
booktitle = { 2018 IEEE/CVF Conference on Computer Vision and Pattern Recognition (CVPR) },
title = {{ ShuffleNet: An Extremely Efficient Convolutional Neural Network for Mobile Devices }},
year = {2018},
volume = {},
ISSN = {},
pages = {6848-6856},
doi = {10.1109/CVPR.2018.00716},
url = {https://doi.ieeecomputersociety.org/10.1109/CVPR.2018.00716},
publisher = {IEEE Computer Society},
address = {Los Alamitos, CA, USA},
month =Jun}

@article{EfficientNet,
  title={EfficientNet: Rethinking Model Scaling for Convolutional Neural Networks},
  author={Mingxing Tan and Quoc V. Le},
  journal={ArXiv},
  year={2019},
  volume={abs/1905.11946},
  url={https://api.semanticscholar.org/CorpusID:167217261}
}

@INPROCEEDINGS {regnet,
author = { Radosavovic, Ilija and Kosaraju, Raj Prateek and Girshick, Ross and He, Kaiming and Dollar, Piotr },
booktitle = { 2020 IEEE/CVF Conference on Computer Vision and Pattern Recognition (CVPR) },
title = {{ Designing Network Design Spaces }},
year = {2020},
volume = {},
ISSN = {},
pages = {10425-10433},
doi = {10.1109/CVPR42600.2020.01044},
url = {https://doi.ieeecomputersociety.org/10.1109/CVPR42600.2020.01044},
publisher = {IEEE Computer Society},
address = {Los Alamitos, CA, USA},
month =Jun}

@INPROCEEDINGS {DLA,
author = { Yu, Fisher and Wang, Dequan and Shelhamer, Evan and Darrell, Trevor },
booktitle = { 2018 IEEE/CVF Conference on Computer Vision and Pattern Recognition (CVPR) },
title = {{ Deep Layer Aggregation }},
year = {2018},
volume = {},
ISSN = {},
pages = {2403-2412},
doi = {10.1109/CVPR.2018.00255},
url = {https://doi.ieeecomputersociety.org/10.1109/CVPR.2018.00255},
publisher = {IEEE Computer Society},
address = {Los Alamitos, CA, USA},
month =Jun}

@article{CCT,
  title={Escaping the Big Data Paradigm with Compact Transformers},
  author={Ali Hassani and Steven Walton and Nikhil Shah and Abulikemu Abuduweili and Jiachen Li and Humphrey Shi},
  journal={ArXiv},
  year={2021},
  volume={abs/2104.05704},
  url={https://api.semanticscholar.org/CorpusID:233210459}
}

@INPROCEEDINGS{resnet,
  author={He, Kaiming and Zhang, Xiangyu and Ren, Shaoqing and Sun, Jian},
  booktitle={2016 IEEE Conference on Computer Vision and Pattern Recognition (CVPR)}, 
  title={Deep Residual Learning for Image Recognition}, 
  year={2016},
  volume={},
  number={},
  pages={770-778},
  doi={10.1109/CVPR.2016.90}}

@INPROCEEDINGS{densenet,
  author={Huang, Gao and Liu, Zhuang and Van Der Maaten, Laurens and Weinberger, Kilian Q.},
  booktitle={2017 IEEE Conference on Computer Vision and Pattern Recognition (CVPR)}, 
  title={Densely Connected Convolutional Networks}, 
  year={2017},
  volume={},
  number={},
  pages={2261-2269},
  doi={10.1109/CVPR.2017.243}}

@inproceedings{scaling,
author = {Grigsby, J. Elisenda and Lindsey, Kathryn and Rolnick, David},
title = {Hidden symmetries of ReLU networks},
year = {2023},
publisher = {JMLR.org},
booktitle = {Proceedings of the 40th International Conference on Machine Learning},
articleno = {471},
numpages = {27},
location = {Honolulu, Hawaii, USA},
series = {ICML'23}
}

@inproceedings{redundant,
 author = {Fukumizu, Kenji},
 booktitle = {Advances in Neural Information Processing Systems},
 editor = {D. Touretzky and M.C. Mozer and M. Hasselmo},
 pages = {},
 publisher = {MIT Press},
 title = {Active Learning in Multilayer Perceptrons},
 url = {https://proceedings.neurips.cc/paper_files/paper/1995/file/8248a99e81e752cb9b41da3fc43fbe7f-Paper.pdf},
 volume = {8},
 year = {1995}
}

@inproceedings{adamw,
  author       = {Ilya Loshchilov and
                  Frank Hutter},
  title        = {Decoupled Weight Decay Regularization},
  booktitle    = {7th International Conference on Learning Representations, {ICLR} 2019,
                  New Orleans, LA, USA, May 6-9, 2019},
  publisher    = {OpenReview.net},
  year         = {2019},
  url          = {https://openreview.net/forum?id=Bkg6RiCqY7},
  bibsource    = {dblp computer science bibliography, https://dblp.org}
}

@article{FisherMatrix,
title = {A Regularity Condition of the Information Matrix of a Multilayer Perceptron Network},
journal = {Neural Networks},
volume = {9},
number = {5},
pages = {871-879},
year = {1996},
issn = {0893-6080},
doi = {https://doi.org/10.1016/0893-6080(95)00119-0},
url = {https://www.sciencedirect.com/science/article/pii/0893608095001190},
author = {Kenji Fukumizu}
}

\appendix
\onecolumn

\section{Empirical Validation of Equation~\ref{eq:approx_mean_var}  }
\label{appendix:meanvar}

\begin{figure}[htbp]
\centering
\includegraphics[width=1\textwidth]{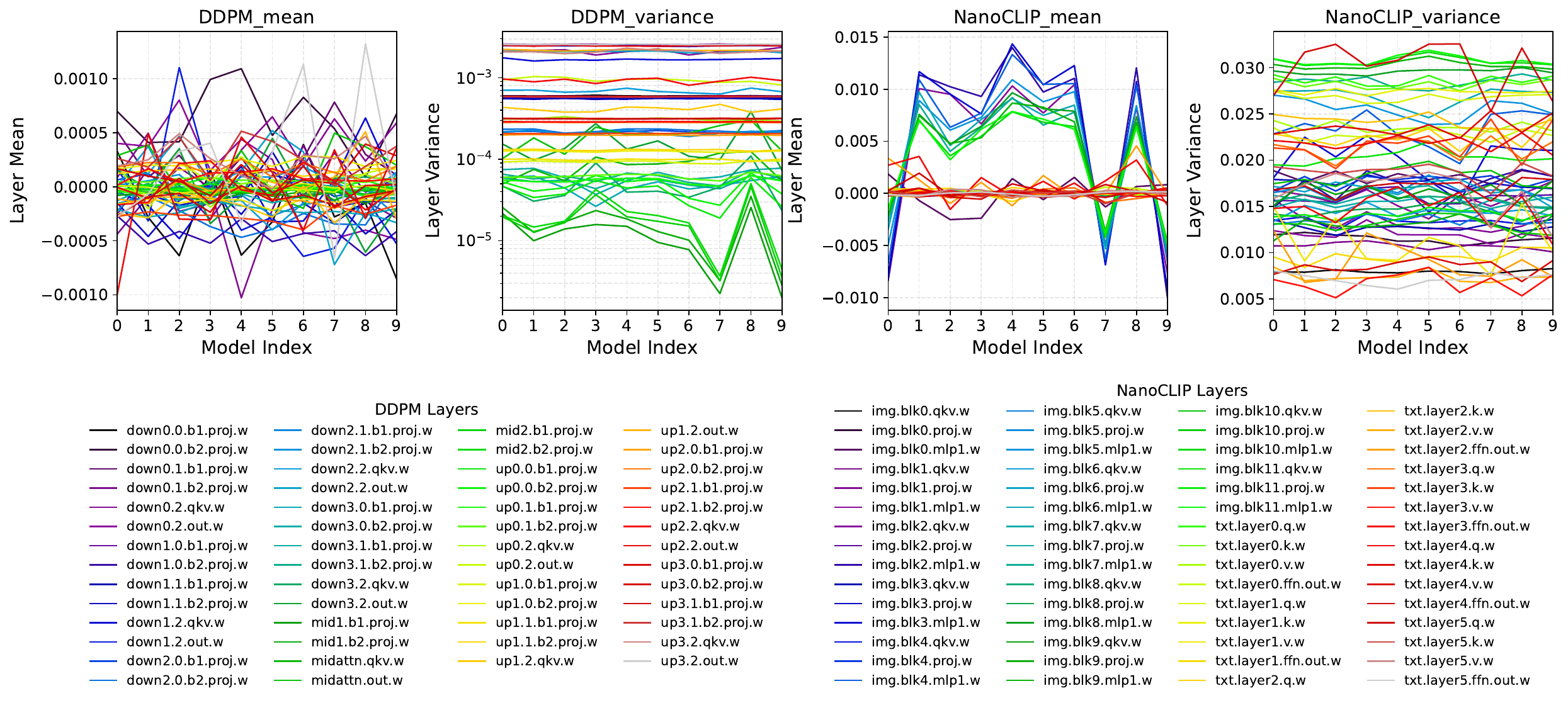}
\caption{
Layer-wise weights mean and variance across independently trained NanoCLIP on Flickr30k and DDPM on Flowers102 models with different random seeds. The x-axis denotes the model index, and the y-axis represents the mean or variance of layer parameters. Each curve corresponds to one layer following the PyTorch naming convention. Batch normalization layers are excluded, as their statistics are highly sensitive to the training batches. The results show that layer-wise variances remain consistent across independent training runs and layer-wise means remain centered around zero.}
\label{fig:meanvar}
\end{figure}

\FloatBarrier

\section{Mode-Connection Continuity Check via Linear Interpolation}
\label{appendix:continuity}



\begin{figure}[!htbp]
    \centering
    \includegraphics[width=0.99\columnwidth]{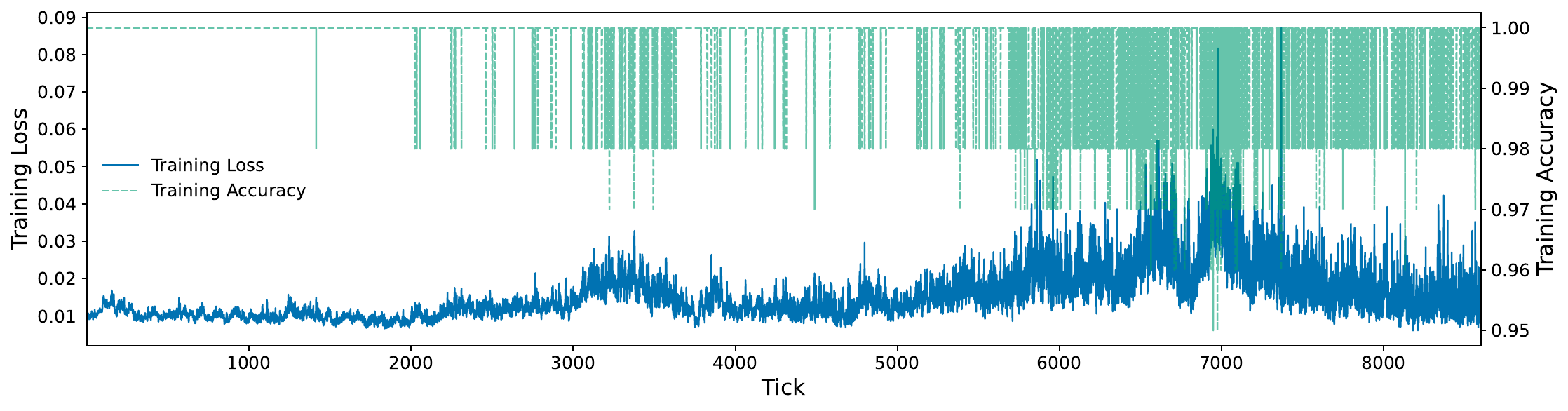}
    \caption{Mode-connection continuity check for NanoCLIP. Twenty uniformly spaced interpolation points are inserted between every pair of consecutive optimization ticks. All interpolation points also remain within the low-loss region.}
    \label{fig:nanoclip_continuity}
\end{figure}

Figure~\ref{fig:nanoclip_continuity} presents the resulting loss trajectory. Despite the substantially denser sampling, no observable loss barrier appears along the constructed path. The maximum training loss increases by less than 5.3\%. The result indicates that the interpolated models remain within the same low-loss region as the optimization trajectory.

\FloatBarrier

\section{Hyperparameters and Configurations of Architecture-aware Connection Building Algorithm}
\label{appendix:hyperparameters}

\begin{table}[p]
\centering
\resizebox{0.92\textwidth}{!}{
\begin{tabular}{cccccc}
\toprule
Stage &
Layer Moving Sequence $\mathcal{U}$ &
Iteration $T$ &
\makecell[c]{
Training Refinement\\
Round $r$
} &
\makecell[c]{
Optimizer and\\
Hyperparameter
} &
\makecell[c]{
Step-size Hyperparameters\\
($\alpha,\beta,\gamma$)
}
\\
\midrule
\multicolumn{3}{c}{Model: DDPM-Flowers102} &
\multicolumn{3}{c}{Runtime: 59 hours (NVIDIA RTX 5090)}
\\
\midrule
1&
init\_conv.\{weight+bias\}
&
400
&
\multirow{39}{*}{
\makecell[c]{
Train until\\
loss$<0.03$\\
or rounds$>500$
}
}
&
\multirow{39}{*}{
\makecell[c]{
AdamW\\
(lr=$1e^{-6}$, wd=0)
}
}
&
\multirow{39}{*}{
(0.001, 0.002, 0)
}
\\
2&
Stage 1 + mlp + downs.0.0
&400&&&\\
3&
Stage 2 + downs.0.1
&400&&&\\
4&
Stage 3 + downs.0.2
&400&&&\\
5&
Stage 4 + downs.0.3
&400&&&\\
6&
Stage 5 + ups.3.0
&400&&&\\
7&
Stage 6 + ups.3.1
&400&&&\\
8&
Stage 7 + ups.3.2
&400&&&\\
9&
Stage 8 + ups.3.3
&400&&&\\
10&
Stage 9 + final\_res\_block
&400&&&\\
11&
Stage 10 + final\_conv
&400&&&\\
12&
Stage 11 + downs.1.0
&400&&&\\
13&
Stage 12 + downs.1.1
&400&&&\\
14&
Stage 13 + downs.1.2
&400&&&\\
15&
Stage 14 + downs.1.3
&400&&&\\
16&
Stage 15 + ups.2.0
&400&&&\\
17&
Stage 16 + ups.2.1
&400&&&\\
18&
Stage 17 + ups.2.2
&400&&&\\
19&
Stage 18 + ups.2.3
&400&&&\\
20&
Stage 19 + downs.2.0
&400&&&\\
21&
Stage 20 + downs.2.1
&400&&&\\
22&
Stage 21 + downs.2.2
&400&&&\\
23&
Stage 22 + downs.2.3
&400&&&\\
24&
Stage 23 + ups.1.0
&400&&&\\
25&
Stage 24 + ups.1.1
&400&&&\\
26&
Stage 25 + ups.1.2
&400&&&\\
27&
Stage 26 + ups.1.3
&400&&&\\
28&
Stage 27 + downs.3.0
&400&&&\\
29&
Stage 28 + downs.3.1
&400&&&\\
30&
Stage 29 + downs.3.2
&400&&&\\
31&
Stage 30 + downs.3.3
&400&&&\\
32&
Stage 31 + mid\_block1
&400&&&\\
33&
Stage 32 + mid\_attn
&400&&&\\
34&
Stage 33 + mid\_block2
&400&&&\\
35&
Stage 34 + ups.0.0
&400&&&\\
36&
Stage 35 + ups.0.1
&400&&&\\
37&
Stage 36 + ups.0.2
&400&&&\\
38&
Stage 37 + ups.0.3
&400&&&\\
39&
All layers
&1200&&&\\

\midrule
\multicolumn{3}{c}{Model: NanoCLIP-Flickr30k} &
\multicolumn{3}{c}{Runtime: 77 hours (NVIDIA RTX 5090)}
\\
\midrule
1&
\makecell[c]{
txt.\{embeddings+pooler+fc\}\\
img.\{patch\_embed+cls+pos+mask+fc\}
}
&
400
&
\multirow{20}{*}{
\makecell[c]{
Train until\\
loss$<0.04$\\
or rounds$>200$
}
}
&
\multirow{20}{*}{
\makecell[c]{
AdamW\\
(lr=$1e^{-4}$, wd=0)
}
}
&
\multirow{20}{*}{
(0.005, 0, 0)
}
\\
2&
Stage 1 + blocks.0
&400&&&\\
3&
Stage 2 + blocks.1
&400&&&\\
4&
Stage 3 + layer.0
&400&&&\\
5&
Stage 4 + blocks.2
&400&&&\\
6&
Stage 5 + blocks.3
&400&&&\\
7&
Stage 6 + layer.1
&400&&&\\
8&
Stage 7 + blocks.4
&400&&&\\
9&
Stage 8 + blocks.5
&400&&&\\
10&
Stage 9 + layer.2
&400&&&\\
11&
Stage 10 + blocks.6
&400&&&\\
12&
Stage 11 + blocks.7
&400&&&\\
13&
Stage 12 + layer.3
&400&&&\\
14&
Stage 13 + blocks.8
&400&&&\\
15&
Stage 14 + blocks.9
&400&&&\\
16&
Stage 15 + layer.4
&400&&&\\
17&
Stage 16 + blocks.10
&400&&&\\
18&
Stage 17 + blocks.11
&400&&&\\
19&
Stage 18 + layer.5
&400&&&\\
20&
All layers
&1000&&&\\
\bottomrule
\end{tabular}
}
\caption{
Hyperparameters used for the architecture-aware connection building algorithm. For each model, the whole process is divided into multiple stages, where only the parameter subsets specified in the layer moving sequence are optimized in each stage. The layer names are abbreviated in the table, following the PyTorch implementation of the corresponding architecture. We also report the total runtime for constructing the complete low-loss path.
}
\label{tab:Hyperparameters}
\end{table}

\FloatBarrier

\section{Hyperparameters and Configurations of AutoNEB}
\label{appendix:autoneb_config}

\begin{table}[!htbp]
\centering
{\small
\setlength{\tabcolsep}{4mm}
\begin{tabular}{cccc}

\toprule
Cycle &
\makecell[c]{
Updates per Pivot
} &
\makecell[c]{
Trainable Internal Pivots
} &
\makecell[c]{
Optimizer and Hyperparameter
}
\\

\midrule
\multicolumn{2}{c}{Model: DDPM-Flowers102} &
\multicolumn{2}{c}{Runtime: 23 hours (NVIDIA A6000)}
\\

\multicolumn{2}{c}{Model: NanoCLIP-Flickr30k} &
\multicolumn{2}{c}{Runtime: 41 hours (NVIDIA A6000)}
\\

\midrule
1 & 1000 & 1 &
\multirow{6}{*}{
\makecell[c]{
AdamW\\
(lr=$1e^{-3}$, wd=0)
}
}
\\
2 & 1000 & 3 & \\
3 & 1000 & 7 & \\
4 & 1000 & 11 & \\
5 & 2000 & 11 & \\
6 & 2000 & 11 & \\

\midrule
7 & 1000 & 11 &
\multirow{4}{*}{
\makecell[c]{
AdamW\\
(lr=$1e^{-4}$, wd=0)
}
}
\\
8  & 1000 & 11 & \\
9  & 1000 & 11 & \\
10 & 1000 & 11 & \\

\midrule
11 & 1000 & 11 &
\multirow{4}{*}{
\makecell[c]{
AdamW\\
(lr=$1e^{-5}$, wd=0)
}
}
\\
12 & 1000 & 11 & \\
13 & 1000 & 11 & \\
14 & 1000 & 11 & \\

\bottomrule
\end{tabular}
}
\caption{
Hyperparameters used for the AutoNEB baseline. Both DDPM and NanoCLIP use the same 14-cycle optimization schedule. Updates per pivot denotes the number of AdamW updates applied to each trainable internal pivot in a cycle. The two endpoint pivots remain fixed and are excluded from the pivot counts and the final path contains 13 pivots in total. AdamW states are reset at the beginning of each cycle. We also report the total runtime.
}
\label{tab:AutoNEBHyperparameters}
\end{table}

\section{Experimental Results of Endpoint Parameter Proximity and Linear Interpolation }
\label{appendix:barrier}

\begin{figure}[htbp]
    \centering    \includegraphics[width=0.99\columnwidth]{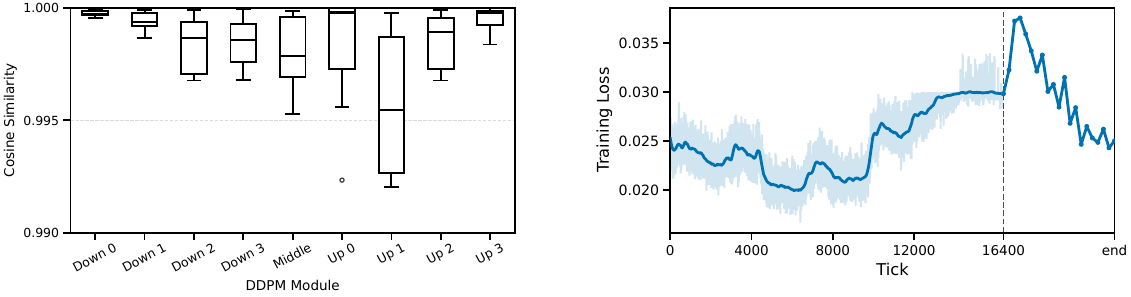}
    \caption{Parameter-space proximity and training loss near the DDPM end mode. \textit{Left}: layer-wise cosine similarity between the DDPM checkpoint at tick 16400 and the end mode. Each box summarizes the distribution of cosine similarity values for multiple layers within one DDPM module, indicating they are highly close in parameter space. \textit{Right}: training loss near the end mode. The left part shows the training loss along the original DDPM low-loss path. The right part shows the training loss obtained by linearly interpolating between the checkpoint at tick 16400 and the end mode using uniformly sampled interpolation points.}
    \label{fig:ddpm_barrier}
\end{figure}

\end{document}